\documentclass[default]{sn-jnl}
\usepackage{setspace}
\usepackage{graphicx}
\usepackage{multirow}
\usepackage{amsmath,amssymb,amsfonts}
\usepackage{amsthm}
\usepackage{mathrsfs}
\usepackage[title]{appendix}
\usepackage{xcolor}
\usepackage{textcomp}
\usepackage{manyfoot}
\usepackage{booktabs}
\usepackage{natbib}
\usepackage{algorithm2e}
\usepackage{algorithmicx}
\usepackage{algpseudocode}
\usepackage{listings}
\usepackage{tabularx}
\usepackage{tabulary}
\usepackage{longtable}
\usepackage{siunitx}
\usepackage{makecell}
\usepackage{subcaption}
\usepackage{url}

\begin{document}

\title[]{Improved Cotton Leaf Disease Classification Using Parameter-Efficient Deep Learning Framework}

\author*[1]{\fnm{Aswini Kumar} \sur{Patra}}\email{aswinipatra@gmail.com}

\author[2]{\fnm{Tejashwini} \sur{Gajurel}}\email{gajureltejashwini@gmail.com}



\abstract{Cotton crops, often called "white gold," face significant production challenges, primarily due to various leaf-affecting diseases. As a major global source of fiber, timely and accurate disease identification is crucial to ensure optimal yields and maintain crop health. While deep learning and machine learning techniques have been explored to address this challenge, there remains a gap in developing lightweight models with fewer parameters which could be computationally effective for agricultural practitioners. To address this, we propose an innovative deep learning framework integrating a subset of trainable layers from MobileNet, transfer learning, data augmentation, a learning rate decay schedule, model checkpoints, and early stopping mechanisms. Our model demonstrates exceptional performance, accurately classifying seven cotton disease types with an overall accuracy of 98.42\% and class-wise precision ranging from 96\% to 100\%. This results in significantly enhanced efficiency, surpassing recent approaches in accuracy and model complexity. The existing models in the literature have yet to attain such high accuracy, even when tested on data sets with fewer disease types.
The substantial performance improvement, combined with the lightweight nature of the model, makes it practically suitable for real-world applications in smart farming. By offering a high-performing and efficient solution, our framework can potentially address challenges in cotton cultivation, contributing to sustainable agricultural practices.}

\keywords{Deep Learning, Cotton Leaf Disease, Precision Agriculture, Smart Farming, Transfer Learning.}



\maketitle

\section{Introduction}

Cotton, a vital crop for fiber production globally, often faces significant threats from leaf-borne diseases \cite{meyer2023world} \cite{manavalan_towards_2022}. These diseases can severely impact crop yield and quality, posing challenges to sustainable production. Recent advancements in cotton leaf disease classification have been driven by the adoption of deep learning methods, particularly Convolutional Neural Networks (CNNs). These architectures effectively leverage spatial correlations in images to identify and classify disease symptoms on cotton leaves \cite{nazeer_detection_2024} \cite{m_deep_2021} \cite{manavalan_towards_2022} \cite{pandian_j_improved_2022}.
In addition to deep learning, traditional machine learning methods such as Random Forest (RF), Support Vector Machines (SVM), Naive Bayes, Multi-Layer Perceptron (MLP), k-Nearest Neighbors (k-NN), and Adaptive Boosting (AdaBoost) have been utilized. These approaches rely on curated features like color and texture to classify diseases \cite{patil_perspective_2021}. Also, ensemble methods have demonstrated high effectiveness in classification tasks. For example, a recent study combined the strengths of GoogleNet, AlexNet, VGG-16, and Inception-V3 to classify images as healthy or diseased. This approach employed Continuous Wavelet Transform (CWT) and Fast Fourier Transform (FFT) to preprocess images before feeding them into deep learning models \cite{shahid_ensemble_2024}.
Innovative approaches, including few-shot learning and metric learning, have also shown promising results in cotton disease classification. These methods enable accurate classification even with limited labeled data, addressing challenges related to data scarcity \cite{liang_few-shot_2021}.

The evolution from binary to multi-class classification reflects the increasing complexity of identifying and assessing cotton leaf diseases. Initially, binary classification focused on distinguishing healthy leaves from those affected by a specific disease, such as cotton root rot disease \cite{wang_automatic_2020}. However, the need for multi-class classification emerged to identify a broader range of diseases \cite{islam_deep_2023} \cite{amin_explainable_2022}, addressing the challenge of differentiating between diseases with similar symptoms. Recently, a comprehensive dataset featuring seven disease classes was introduced, marking a significant advancement in classification research \cite{bishshash_compre_2024}.
Beyond standard classification, recent studies have delved into more specialized problems, such as evaluating the susceptibility of plants to curl virus disease \cite{nazeer_detection_2024}. This approach categorizes plants into varying levels of susceptibility, offering a more nuanced understanding of disease progression and plant resistance.

Most research in this domain relies on deep learning models with millions of parameters. However, a growing shift is toward using lightweight deep-learning models that reduce computational overhead. Minimizing the number of parameters makes these models suitable for deployment on resource-limited devices like mobile phones. A recent study emphasizes parameter pruning as a key technique to create smaller, more efficient models without compromising accuracy\cite{zhu_cotton_2022}. This strategy has relevance for practical applications, especially in situations requiring real-time predictions to ensure timely and effective disease management.

Despite advancements in deep learning and traditional machine learning methods, a gap remains in developing models that balance high performance with computational efficiency, particularly for real-time disease management predictions on resource-constrained devices.
This work addresses this gap by leveraging transfer learning, enabling the model to benefit from pre-trained knowledge while focusing on domain-specific cotton disease classification. The proposed pipeline reduces computational complexity by combining transfer learning with a lightweight model and selecting only a few trainable layers while maintaining high classification accuracy.

\section{Materials and Methods}
\subsection{Data Description}
We utilized a publicly available dataset \cite{bishshash_compre_2024} featuring images of cotton leaf diseases captured during field surveys at the National Cotton Research Institute in Gazipur between October 2023 and January 2024. The images were taken with a Redmi Note 11s smartphone under diverse environmental conditions, resulting in varied resolutions of 3000×4000, 2239×2239, and 1597×1597 pixels to represent disease symptoms comprehensively.
The data set comprises seven categories of cotton leaf conditions: bacterial blight, curl virus, herbicide growth damage, leaf hopper jassids, leaf reddening, leaf variegation, and healthy leaves. Two cotton varieties are featured: American Upland (Gossypium hirsutum) and CB-12 to CB-18. American Upland cotton is widely valued for its superior fiber quality and high yield, while CB-12 to CB-18 varieties exhibit distinct agronomic characteristics such as boll count, yield potential, and ginning outturn.

\begin{figure}[h!]
    \centering
    \begin{subfigure}{0.23\textwidth}
        \centering
        \includegraphics[width=\textwidth]{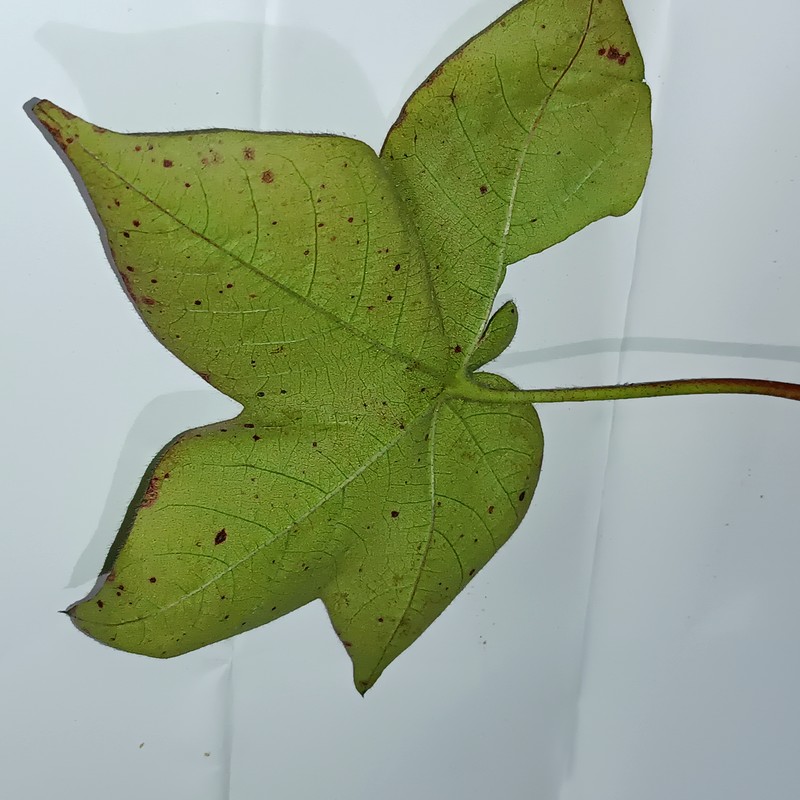}
        \caption{Bacterial Blight}
        \label{fig:sub1}
    \end{subfigure}
    \hfill
    \begin{subfigure}{0.23\textwidth}
        \centering
        \includegraphics[width=\textwidth]{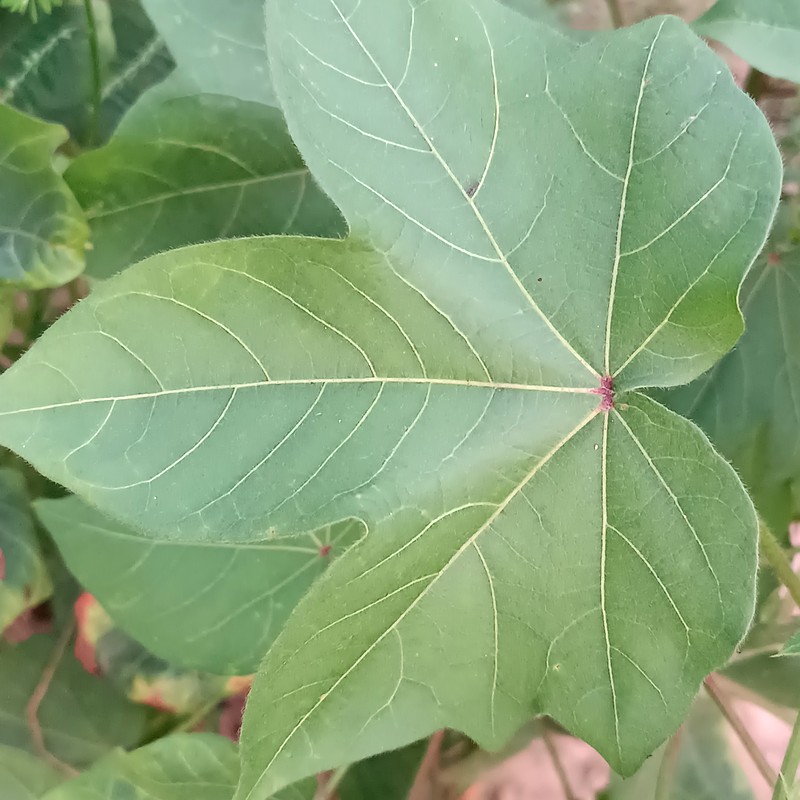}
        \caption{Curl Virus}
        \label{fig:sub2}
    \end{subfigure}
    \hfill
    \begin{subfigure}{0.23\textwidth}
        \centering
        \includegraphics[width=\textwidth]{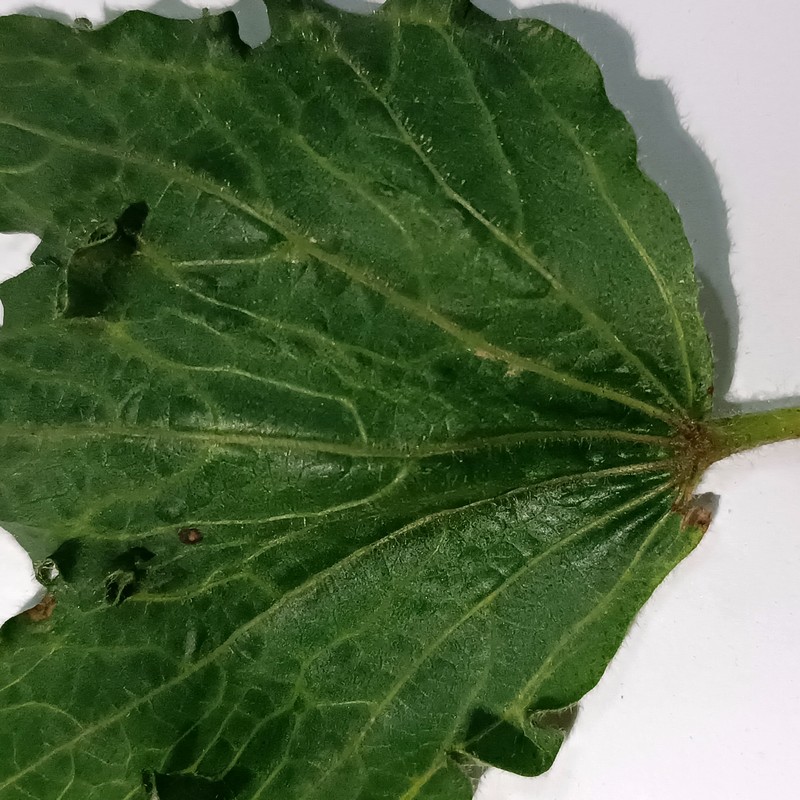}
        \caption{Herbicide Growth Damage}
        \label{fig:sub3}
    \end{subfigure}
    \hfill
    \begin{subfigure}{0.23\textwidth}
        \centering
        \includegraphics[width=\textwidth]{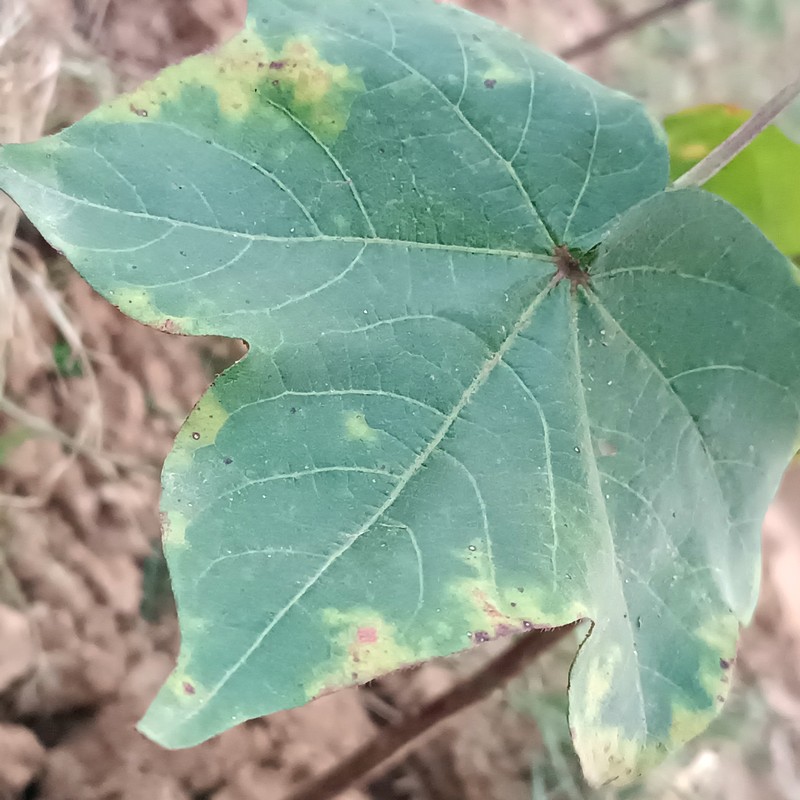}
        \caption{Leaf Hopper Jassids}
        \label{fig:sub4}
    \end{subfigure}
    
    \vspace{0.5cm} 

    \begin{subfigure}{0.23\textwidth}
        \centering
        \includegraphics[width=\textwidth]{Leaf Redding.jpg}
        \caption{Leaf Redding}
        \label{fig:sub5}
    \end{subfigure}
    \hfill
    \begin{subfigure}{0.23\textwidth}
        \centering
        \includegraphics[width=\textwidth]{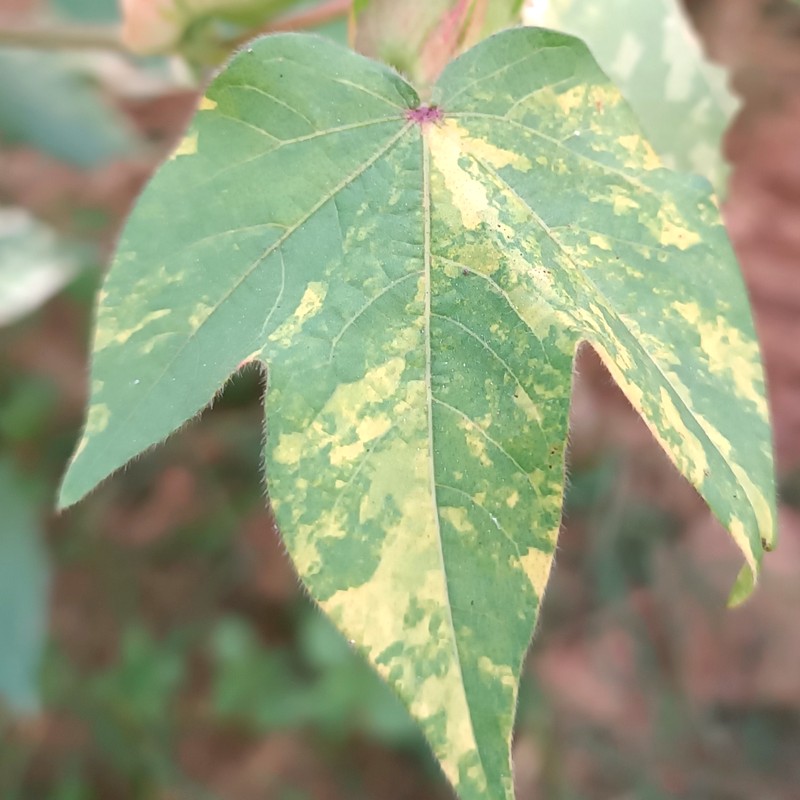}
        \caption{Leaf Variegation}
        \label{fig:sub6}
    \end{subfigure}
    \hfill
    \begin{subfigure}{0.23\textwidth}
        \centering
        \includegraphics[width=\textwidth]{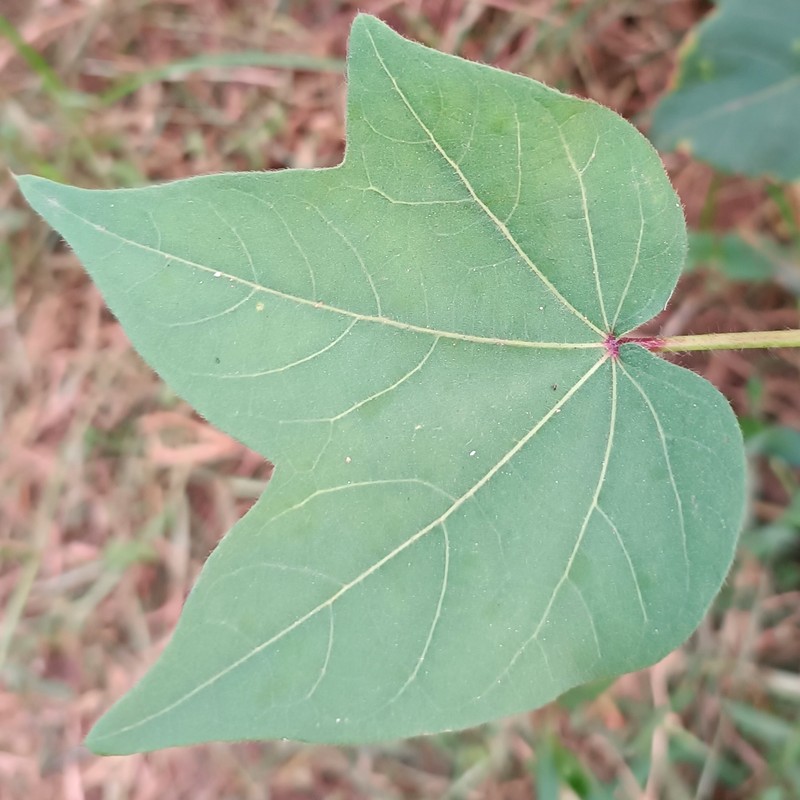}
        \caption{Healthy}
        \label{fig:sub7}
    \end{subfigure}
    \caption{Cotton Leaf Disease Classes}
    \label{fig:seven_subfigures}
\end{figure}

There are 7,000 augmented images in total, with 1,000 images representing each of the seven conditions. The Fig. \ref{fig:seven_subfigures} displays all seven conditions. These categories showcase diverse visual traits, including discoloration, deformation, lesions, and pest infestations.

\subsection{Deep Learning Framework} 
\begin{figure}
\centerline{\includegraphics[scale=0.5]{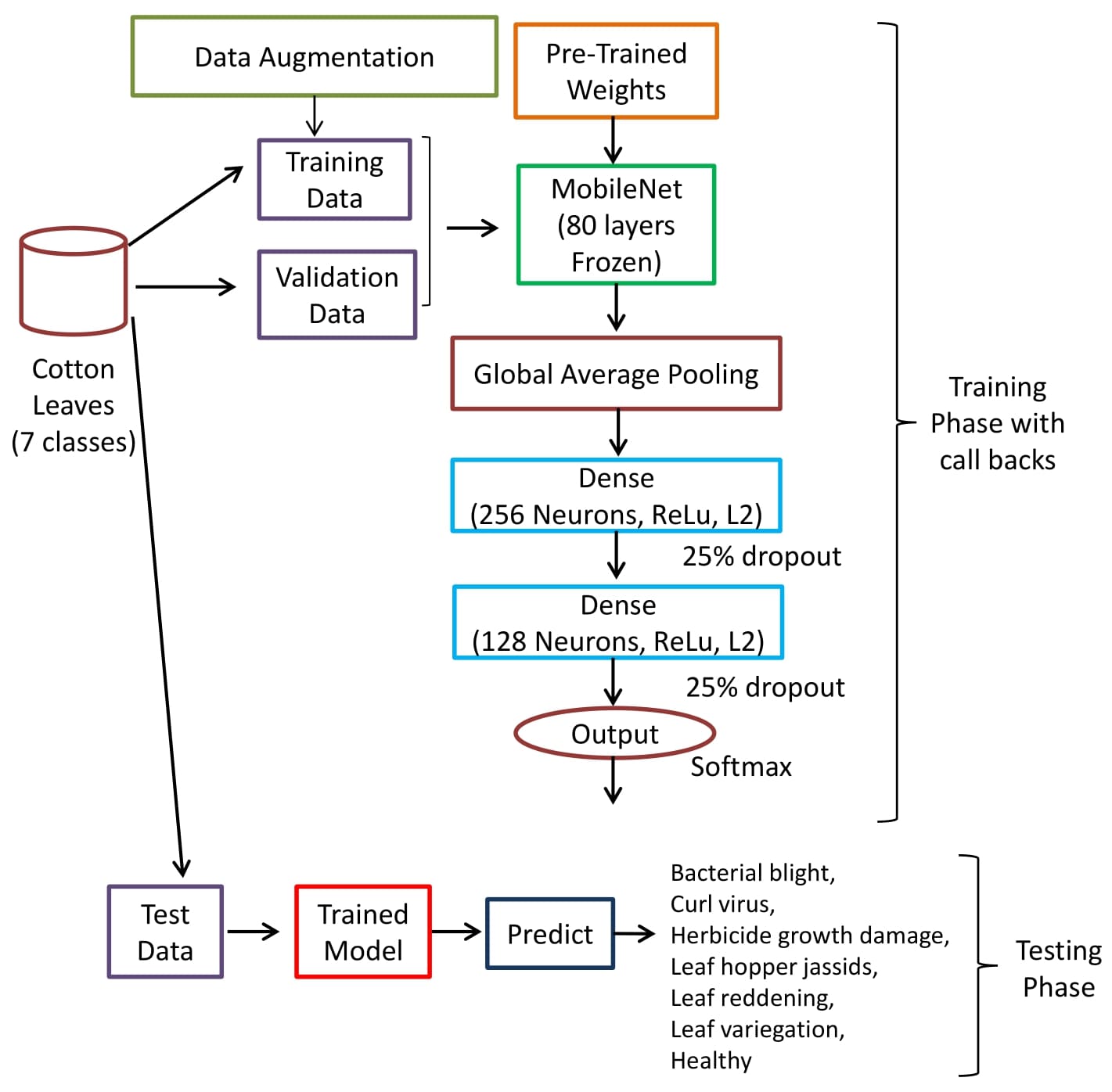}}
\caption{Deep Learning Framework to Predict Cotton Leave Disease}
\label{fig:frame}
\end{figure}

The framework is built on the CNN based architecture \textit{MobileNet} \cite{howard_mobilenets_2017} with pre-trained weights from the \textit{ImageNet} dataset, leveraging transfer learning to utilize the general visual features learned from the large-scale dataset. The first 80 layers of \textit{MobileNet} are frozen to retain essential feature extraction capabilities, while custom layers are added on top to adapt the model for the specific classification task. The additional layers include a \textit{GlobalAveragePooling2D} layer to reduce spatial dimensions, followed by \textit{dense layers} with 256 and 128 neurons, respectively, using \textit{ReLU} activation and \textit{L2} regularization to mitigate over-fitting. \textit{Dropout} layers with a 25\% rate are incorporated for additional regularization. The final output layer uses a \textit{softmax} activation to predict the class probabilities corresponding to the number of categories in the data set.
The model employs data augmentation during training to increase diversity in the dataset and improve robustness. Callbacks such as model checkpoint, early stopping and learning rate reduction are used to improve training efficiency and avoid over-fitting. Thus, this approach effectively combines the feature extraction strength of a pre-trained backbone with task-specific adaptations, ensuring optimal performance in classifying cotton leaf diseases.
The framework of the model is illustrated in Fig. \ref{fig:frame}.

\section{Results and Discussion}
\subsection{Experiment and Performance Evaluation}
We experimented with the proposed model in a Kaggle notebook utilizing the standard CPU environment without any special hardware acceleration. The code was implemented using Python version 3.10.14, leveraging built-in functionalities from libraries such as \textit{Keras}, \textit{TensorFlow}, \textit{Scikit-learn}, \textit{Pandas}, \textit{NumPy}, and \textit{Matplotlib}.

The dataset is initially split into 80\% for training and 20\% for validation and testing, with a \textit{random\_state} of 42 to ensure reproducibility. The 20\% validation and test set is then further divided into separate validation and testing subsets, again using a \textit{random\_state} of 42. For the data preprocessing, data generators are used for all three sets (training, validation, and testing). In the training set, eight augmentation techniques are applied, namely rescaling, rotation, width, height shifting, shear transformation, zooming, and horizontal flipping. These augmentations help increase the model's robustness by introducing variety into the training data. For the validation and test sets, only rescaling is applied to preserve the integrity of the evaluation process without introducing any additional variations. The model is trained for 40 epochs using the \textit{AdamW} optimizer and \textit{categorical cross-entropy} loss function. The image input size for the model is set to 224x224 pixels. The parameters used in the training of the model are listed in Table \ref{tab:cnn-parameters}.

\begin{table}[ht]
    \centering
    \caption{Parameters for Model Training}
    \begin{tabular}{|c|c|}
        \hline
        \textbf{Parameter Type} & \textbf{Parameters and Values} \\ \hline
        Data Augmentation & 
        \begin{tabular}[c]{@{}l@{}}
        rescale=\( \frac{1}{255} \) \\ 
        rotation range=20 \\ 
        width shift range=0.2 \\ 
        height shift range=0.2 \\ 
        shear range=0.2 \\ 
        zoom range=0.2 \\ 
        horizontal flip=True
        \end{tabular} \\ \hline
        Model Configuration & 
        \begin{tabular}[c]{@{}l@{}}
        optimizer=AdamW \\ 
        loss function=categorical\_cross\_entropy \\ 
        Image Size=(224, 224) \\ 
        Batch Size=32 \\ 
        Epochs=40
        \end{tabular} \\ \hline
    \end{tabular}
    \label{tab:cnn-parameters}
\end{table}

\begin{table}[ht]
    \centering
    \caption{Callback Parameters for Model Training}
    \begin{tabular}{|c|c|}
        \hline
        \textbf{Callback} & \textbf{Parameters and Values} \\ \hline
        ModelCheckpoint & 
        \begin{tabular}[c]{@{}l@{}}
        save\_best\_only=True \\ 
        monitor='val\_loss' \\ 
        mode='min'
        \end{tabular} \\ \hline
        EarlyStopping & 
        \begin{tabular}[c]{@{}l@{}}
        monitor='val\_loss' \\ 
        patience=10 \\ 
        restore\_best\_weights=True
        \end{tabular} \\ \hline
        ReduceLROnPlateau & 
        \begin{tabular}[c]{@{}l@{}}
        monitor='val\_loss' \\ 
        factor=0.5 \\ 
        patience=5 \\ 
        min\_lr=1e-6
        \end{tabular} \\ \hline
        \multicolumn{2}{|c|}{Initial Learning Rate = 0.001} \\ \hline
    \end{tabular}
    \label{tab:callbacks}
\end{table}

 For the training, the model uses three callbacks: \textit{ModelCheckpoint}, \textit{EarlyStopping}, and \textit{ReduceLROnPlateau}. Table \ref{tab:callbacks} outlines the key parameters used with callbacks. The initial learning rate for the training optimizer is set at 0.001.
The \textit{ModelCheckpoint} callback ensures that the model saves its weights only when the validation loss improves, preventing unnecessary storage of suboptimal models. The monitoring metric is validation loss (val\_loss), and the mode is set to min to indicate that lower values are better.
The \textit{EarlyStopping} callback halts training if the validation loss does not improve after 10 consecutive epochs. Additionally, the best weights from training are restored, ensuring the final model is balanced.
Lastly, the \textit{ReduceLROnPlateau} callback adjusts the learning rate dynamically when the validation loss plateaus. If there is no improvement in val\_loss for 5 epochs, the learning rate is halved (factor = 0.5). However, the learning rate will not exceed a minimum threshold of 1e-6.
These callback configurations are essential for efficient model training, allowing dynamic adjustments and early termination to prevent over-fitting and optimize performance.

\begin{figure}[h!]
    \centering
    \begin{subfigure}{0.45\textwidth}
        \centering
        \includegraphics[width=0.95\textwidth]{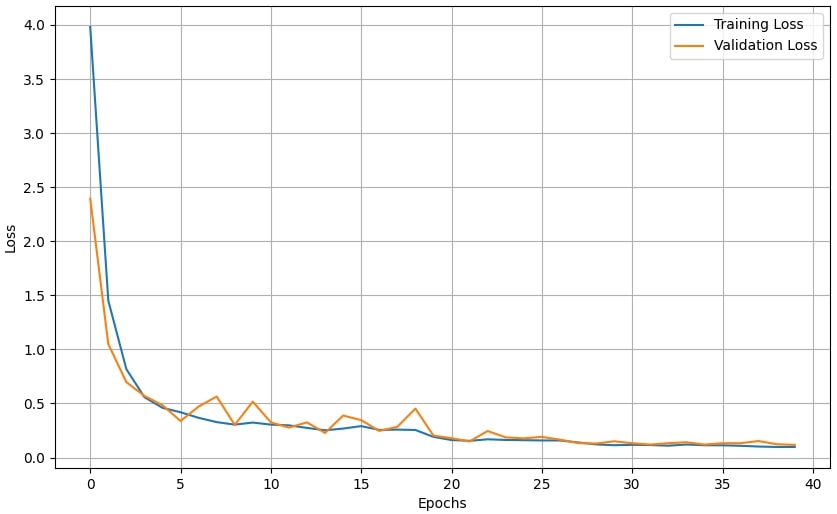}
        \caption{}
        \label{fig:loss}
    \end{subfigure}
    \hfill
    \begin{subfigure}{0.45\textwidth}
        \centering
        \includegraphics[width=0.95\textwidth]{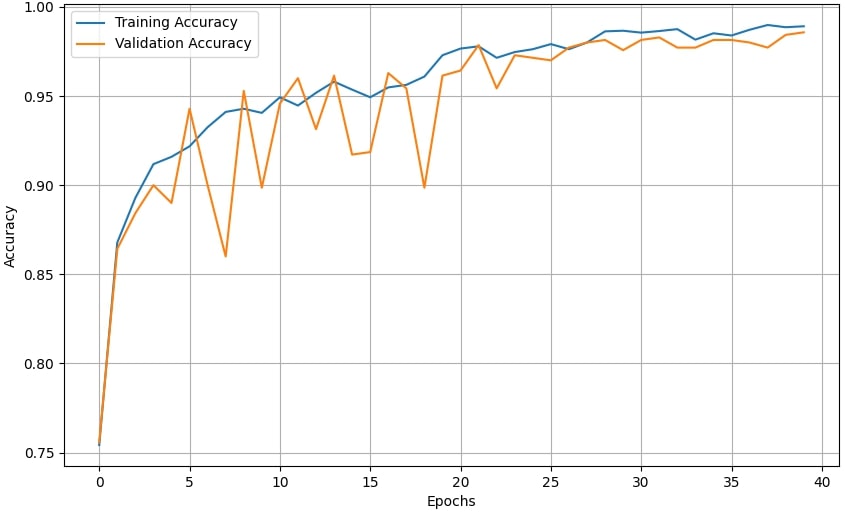}
        \caption{}
        \label{fig:acc}
    \end{subfigure}
    \caption{Learning Curves: a) Training vs Validation Loss, b) Training vs Validation Accuracy}
    \label{fig:curves}
\end{figure}

\begin{figure}[!htp]
\centerline{\includegraphics[scale=0.35]{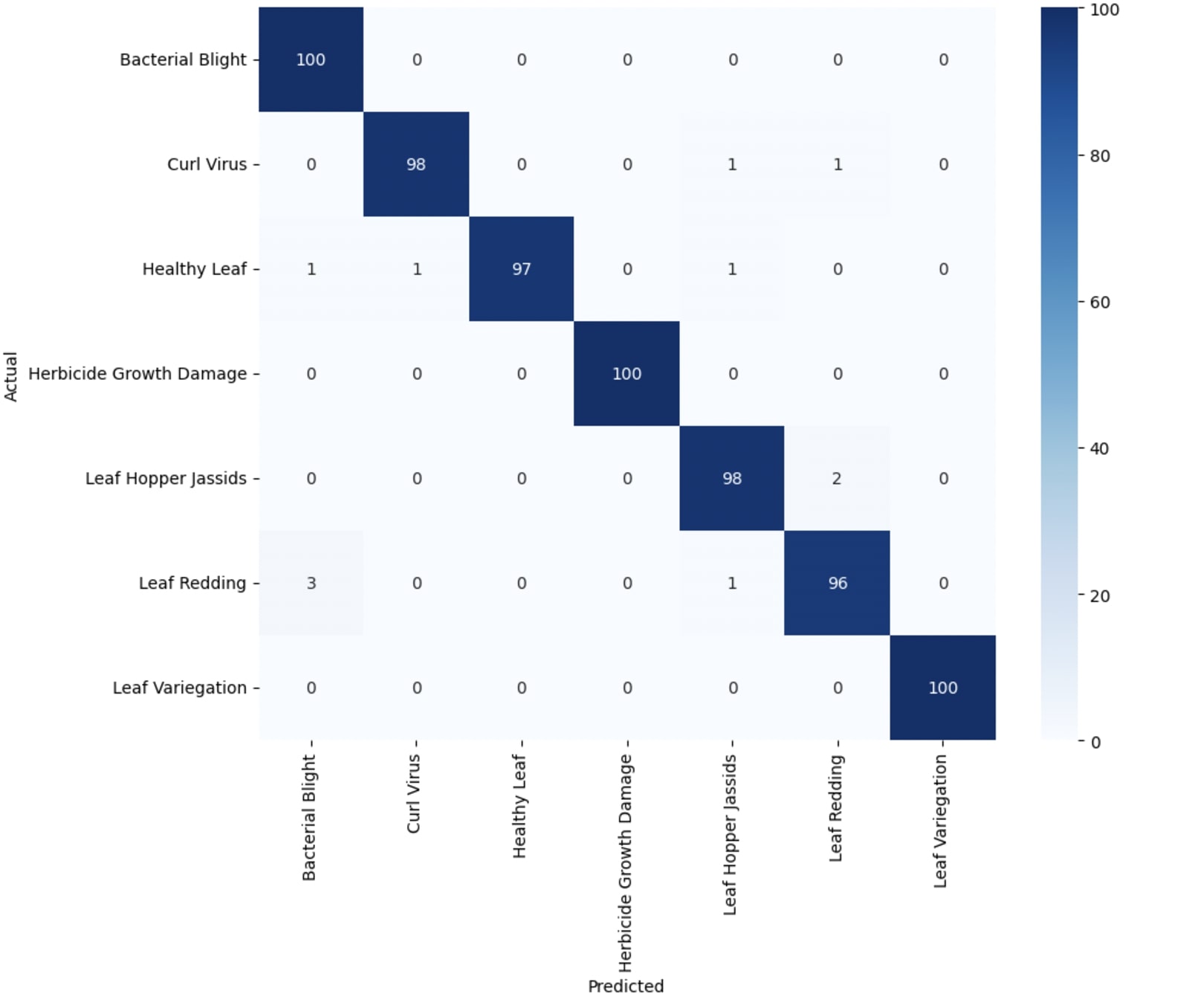}}
\caption{Confusion Matrix}
\label{fig:config}
\end{figure}

During training, the model is evaluated on the validation set at the end of each epoch, providing insights into its performance through metrics such as accuracy and loss. This allows for tracking the model's progress, ensuring that it improves over time, and helps in identifying potential issues such as over-fitting or under-fitting. The learning curves generated from this process are useful for visualizing the model's convergence and generalization ability. Figures \ref{fig:loss} and \ref{fig:acc} show the learning curves for loss and accuracy, respectively. Together, these curves indicate a well-trained model with good generalization. The complementary alignment of the training and validation curves in both loss and accuracy plots suggests that the model is robust and avoids over-fitting.
The gradual and consistent reduction in loss reflects an effective learning process. The steady rise in accuracy further indicates the model's ability to adapt successfully to the training data. At the same time, the model maintains strong performance on unseen validation data. The model was saved at the 37th epoch based on the defined callback procedures, and the trained model was used to predict the classes of the test images. The diagonal values in the confusion matrix (Fig. \ref{fig:config}) represent correct predictions, with bacterial blight, herbicide growth damage, and leaf variegation achieving 100\% accuracy. Mis-classifications are minimal, as shown in the off-diagonal values: leaf redding, healthy leaf, curl virus, and leaf hopper jassids collectively account for 11 mis-classifications out of 700 tested images. Overall, the model demonstrates strong performance with an accuracy of 98.42\%.

\subsection{Comparative Analysis}
We compared the performance of our proposed framework with an InceptionV3-based approach using the same dataset, evaluating metrics such as precision, recall, F1-score, accuracy, and the number of parameters. The results are presented in Table \ref{tab:performance_comparison}. Our framework demonstrated notable improvements, achieving higher precision and recall for 6 and 5 disease classes, respectively. Furthermore, the F1 score surpassed the InceptionV3-based method in 6 classes and matched it in one class. While the overall accuracy of our framework was 89.42\% compared to 96.03\% for the InceptionV3-based approach, our model excelled in class-level metrics, showcasing its strength in disease-specific classification. Moreover, our model achieves higher accuracy with a sixfold reduction in parameters. Specifically, our framework contains only 3.5 million parameters compared to 23.9 million parameters in the InceptionV3-based method. This lightweight design makes our model highly suitable for mobile-based testing, enabling farmers to use it easily and efficiently.

\begin{table}[!tp]
\centering
\caption{Performance Comparison}
\begin{tabular}{|l|c|c|c|c|c|}
\hline
\rotatebox{270}{\textbf{Class Name}} & \rotatebox{270}{\textbf{Precision}} & \rotatebox{270}{\textbf{Recall}} & \rotatebox{270}{\textbf{F1 Score}}  & \rotatebox{270}{\textbf{Accuracy}} & \rotatebox{270}{\textbf{No. of Parameters}} \\ \hline
\multicolumn{6}{|c|}{\textbf{InceptionV3 Based Framework \cite{bishshash_compre_2024}}} \\ \hline
Bacterial Blight       & 0.89 & 0.96 & 0.92 & \textbf{} & \textbf{} \\ 
Curl Virus             & 0.99 & 0.97 & 0.98 & \textbf{} & \textbf{} \\ 
Healthy Leaf           & 0.94 & 0.98 & 0.96 & \textbf{} & \textbf{} \\ 
Herbicide Growth \newline Damage & 0.98 & 1.00 & 0.98 & 96.03\% & 23.9 \\ 
Leaf Hopper Jassids    & 0.90 & 0.98 & 0.94 & \textbf{} & \textbf{million} \\ 
Leaf Redding           & 0.98 & 0.90 & 0.94 & \textbf{} & \textbf{} \\ 
Leaf Variegation       & 1.00 & 1.00 & 1.00 & \textbf{} & \textbf{} \\ \hline
\multicolumn{6}{|c|}{\textbf{Proposed Work}} \\ \hline
Bacterial Blight       & 0.96 & 1.00 & 0.98 & \textbf{} & \textbf{} \\ 
Curl Virus             & 0.99 & 0.98 & 0.98 & \textbf{} & \textbf{} \\ 
Healthy Leaf           & 1.00 & 0.97 & 0.98 & \textbf{} & \textbf{} \\ 
Herbicide Growth \newline Damage & 1.00 & 1.00 & 1.00 & 98.42\% & 3.5  \\ 
Leaf Hopper Jassids    & 0.97 & 0.98 & 0.98 & \textbf{} & \textbf{million} \\ 
Leaf Redding           & 0.97 & 0.96 & 0.96 & \textbf{} & \textbf{} \\ 
Leaf Variegation       & 1.00 & 1.00 & 1.00 & \textbf{} & \textbf{} \\ \hline
\end{tabular}
\label{tab:performance_comparison}
\end{table}

\subsection{Reduced Parameters Set Up}
The model retains the pre-trained feature extractors by freezing the first 80 layers of MobileNet optimized for capturing general patterns such as edges, textures, and shapes. These foundational features are critical for most image recognition tasks and provide a robust starting point for further training. Freezing these layers retains their pre-trained knowledge and significantly reduces computational demands, as the frozen weights remain unchanged during backpropagation. This allows the model to concentrate its learning on the custom task-specific layers designed to capture the unique characteristics of cotton disease features. 
The trainable parameters, comprising approximately 1.36 million (38.5\% of the total parameters), are concentrated in the unfrozen layers of MobileNet (from layer 81 onward) and the newly added custom layers. These include dense layers, dropout layers, and the final softmax layer. Together, these components are designed to adapt the model for the specific classification task by learning complex patterns unique to the dataset. The trainable layers contribute to task-specific learning while maintaining efficiency by targeting only a subset of the total parameters.
In contrast, the non-trainable parameters, amounting to 2.17 million (61.5\% of the total parameters), remain fixed within the frozen layers of MobileNet. These preserved parameters ensure that the influential and generalizable features acquired during pretraining are retained. By utilizing these frozen parameters, the model effectively leverages MobileNet's established capabilities while fine-tuning the unfrozen layers and task-specific additions. 
\section{Conclusion}
This study presents a parameter-efficient deep-learning framework designed to classify cotton leaf diseases accurately. By utilizing pre-trained MobileNet layers through transfer learning, the model preserves generalizable features while effectively adapting to the unique characteristics of the target dataset. This framework achieves an accuracy of 98.42\% in classifying seven cotton leaf conditions, surpassing existing methods while notably reducing model complexity to enhance computational efficiency. Its lightweight design is ideal for resource-constrained environments, enabling real-time disease diagnosis on low-power devices in agricultural settings. By striking a balance between high accuracy and computational efficiency, this study makes deep learning techniques more accessible to the farming sector. The innovative approach integrates cutting-edge learning methods with practical deployment considerations, ensuring strong performance and real-world applicability.

\section*{Data Availability}
The data used in this study are derived from a publicly available dataset, which can be accessed at \url{https://data.mendeley.com/datasets/b3jy2p6k8w/1}. Additionally, the processed data supporting the findings of this study are available from the corresponding author upon reasonable request.
\bibliographystyle{unsrt}
\bibliography{ref}

\end{document}